\documentclass[sigconf]{acmart}
\AtBeginDocument{%
  }

\setcopyright{acmlicensed}
\copyrightyear{2026}
\acmYear{2026}
\acmDOI{XXXXXXX.XXXXXXX}
\acmConference[EMF@HRI'26]{Errors, Mistakes, and Failures in Humans and Robots}{March 16,
  2026}{Edinburgh, UK}
\acmISBN{978-1-4503-XXXX-X/2018/06}

\begin{document}

\title[Designing for Error Recovery]{Designing for Error Recovery in Human-Robot Interaction}

\author{Christopher D. Wallbridge}
\email{wallbridgec@cardiff.ac.uk}
\orcid{0000-0001-9468-122X}
\affiliation{%
  \institution{Cardiff University}
  %\city{Cardiff}
  \country{United Kingdom}
}

\author{Erwin~Jose Lopez~Pulgarin}
\orcid{0000-0001-9927-6688}
\email{erwin.lopezpulgarin@manchester.ac.uk}
\affiliation{%
  \institution{The University of Manchester}
  \city{Manchester}
  \country{United Kingdom}}

\renewcommand{\shortauthors}{Wallbridge and Pulgarin}

\begin{abstract}
    This position paper looks briefly at the way we attempt to program robotic AI systems. Many AI systems are based on the idea of trying to improve the performance of one individual system to beyond so-called human baselines. However, these systems often look at one shot and one-way decisions, whereas the real world is more continuous and interactive. Humans, however, are often able to recover from and learn from errors - enabling a much higher rate of success. We look at the challenges of building a system that can detect/recover from its own errors, using the example of robotic nuclear gloveboxes as a use case to help illustrate examples. We then go on to talk about simple starting designs. 
\end{abstract}




\maketitle

\section{Introduction}

Autonomous robotic systems promise many potential benefits for the real world. Advances in AI enable further possibilities for real-world deployments. Many of these systems look to areas of safety critical deployment, where we hope to be able to engineer away human error. Even as these systems improve, the complexity of the real world may mean it is impossible for us to achieve a system will perform with no errors 100\% of the time.

If we cannot guarantee performance, at what point does it make sense to be able to say a robot is safe to use? Often when creating AI systems, we attempt to achieve performance levels similar to or exceeding human performance as the gold standard. Human performance is sometimes informally quoted at 95\% for a given task. While performance is task-dependent, this is likely based on tasks like image recognition~\cite{russakovsky2015imagenet}, or handling questions that should be trivial~\cite{zellers2019hellaswag}, where we see these benchmarks of human performance. Many of the tasks we try to solve with AI in robotics are these tasks that on the surface seem trivial, but in fact are much harder to program a robot to do; Moravec's paradox~\cite{moravec1988mind}.

However, is this a realistic measure of human performance? Many of these systems look at a single point in time, and a one-way interaction~\cite{ni2025survey}. Humans, however, are well equipped to recover from and learn from their own mistakes~\cite{mera2022unraveling}. Recovery from an error in a task can be both dependent on the task and the type of the error~\cite{jang2014empirical}, but can also be as successful as the original performance. As such, if we take a simplified example of the chance for the task to originally be unsuccessful at 5\%, and the chance for recovery to be unsuccessful at 5\%, the actual likelihood of failure at a task drops to 0.25\%. This still assumes only one chance at recovery, even though real-world interactions are continuous, often allowing for multiple chances of recovery from multiple sources.

This view on error management aligns with the "Swiss Cheese" model~\cite{reason2000human}. This model views each layer of error prevention as a barrier with holes. When all the holes align, this is when a threat can penetrate all the defences. Therefore, the simplest method to lower the rate of errors causing an issue is to add more layers, reducing the possibility that a threat can penetrate all layers at once.

This paper will discuss some of the challenges present in enabling a robotic system to handle a task/operational error or mistake, in the context of a use case - Nuclear Gloveboxes.

\section{Use Case - Nuclear Glovebox}

\begin{figure}[ht]
    \centering
    \includegraphics[width=\columnwidth]{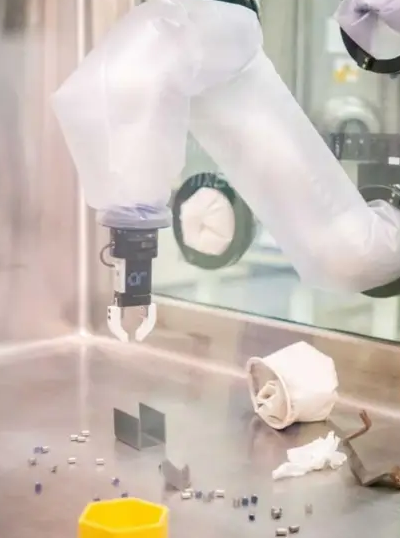}
    \caption[Image showing a robot in a traditional glovebox]{Image showing a robot in a traditional glovebox with a sleeve, part of the RrOBO~\cite{rrobo} project. This is being demonstrated at Sellafield in Cumbria, UK. The robot has been inserted through one of the ports that an operator would normally use to interact with the environment. The sleeve protects the robot overall.}
    \label{fig:Sleeve}
\end{figure}

\begin{figure*}[ht]
    \centering
    \includegraphics[width=\textwidth]{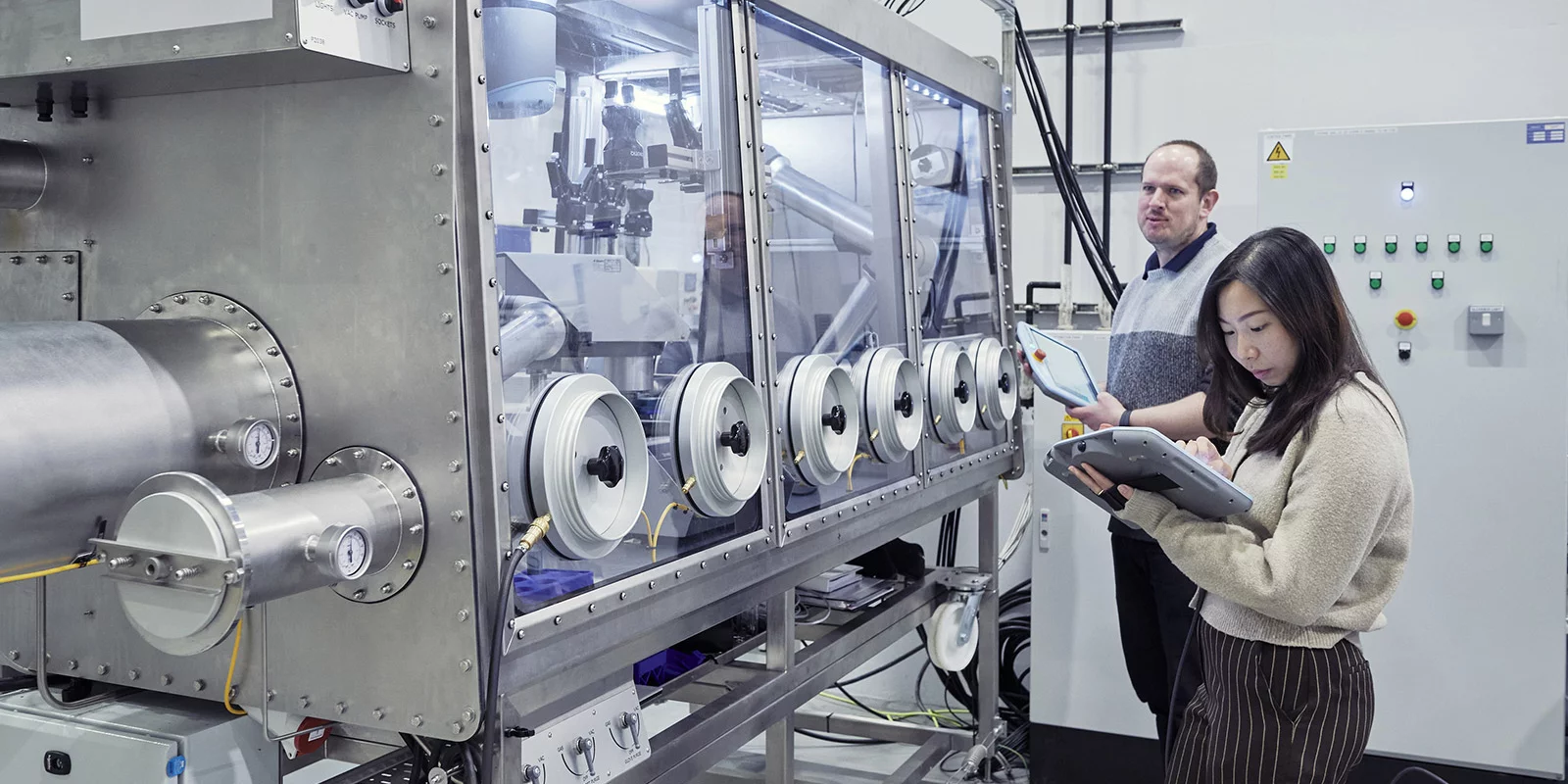}
    \caption[Image showing a purpose built robotic glovebox]{Image showing a purpose built robotic glovebox, part of the RoBox~\cite{robox} project. The system here is being used to develop new methods at the RAICo1 Facility in Cumbria. The robots are built into the glovebox and users are able to control the robot remotely.}
    \label{fig:robot_box}
\end{figure*}

Nuclear decommissioning environments give a prime example of why a system cannot be guaranteed in a real-world task. Nuclear gloveboxes are contained areas used to perform operations on radioactive materials, often waste materials, managed during the decommissioning stage of a facility. Traditionally, users interact with materials inside the glovebox using thick gloves that block radiation. These gloveboxes often offer limited access, are not very ergonomic, and still have risk e.g. if the gloves are pierced. As such, allowing operators to remotely work with the materials using robots can offer many benefits. There are two approaches to achieve this: The first is using a robot in an existing glovebox, the robot is inserted through the usual entry port, and is often sleeved to both prevent as much contamination of the robot as possible, and maintain overall containment (see fig.~\ref{fig:Sleeve}). 
The second is to use a purpose built glovebox with the robots inside, see fig.~\ref{fig:robot_box}). 
The longer a robot is in contact/proximity to radioactive materials, the more likely its performance is to degrade due to radiation damage. Even in these controlled environments, there are often plenty of unknowns~\cite{dennison1995application}. Waste is often being sorted - hazards may be present such as sharp objects, or higher than expected radiation. At the same time, errors in these environments can cause significant issues. These may range from increased task duration -potentially meaning robots and operators spend more time in hazardous environments- to complete equipment degradation -leaving more radioactive waste to be disposed- or in extremes, causing harm to people. Even when small issues occur, if they are not detected, they may compound to cause more severe errors.

\section{Challenges}
\subsection{Defining the Task}

While there is a lot of preparation involved in putting together an operation in a glovebox, there are often unknowns. A common task is inspecting and sorting the contents of a sealed canister, as to understand its level of contamination and decide if it requires processing or not. The canister is opened and its content emptied into the glovebox. Until this point, the exact contents of the canister and its current level of contamination are unknown. Items are usually sorted by how they need to be disposed of - classifications such as the level of radioactivity, and other hazardous properties of the item. In some cases, such item needs to be repacked into a different containment unit or canister, requiring additional manipulation of the items; this includes cutting the item in smaller pieces, i.e. size reduction, or packing it into an additional containment unit such as a plastic bag previous to repacking it into the canister. As such, the task cannot be fully pre-defined, and aspects of the task must be defined during operation. Knowing the goal is key to recognising an error - as it leads away from the goal.

Large Language Models (LLM) have been deployed to attempt to solve such challenges, such as AutoFlow~\cite{li2024autoflow}. However, without proper grounding in the task being performed, this is liable to create LLM "hallucinations"~\cite{xu2024hallucination}. 

\subsection{Identifying Errors}

In any given scenario there are different errors that can occur. Some errors can be defined that are potentially universal across tasks e.g.:
\begin{itemize}
    \item Unexpected Collision.
    \item Failure to/incorrectly grasp an object.
    \item Incorrect identification of an object.
\end{itemize}
Some errors are likely to be specialised or task specific. For instance, looking at our use case of a nuclear glovebox:
\begin{itemize}
    \item Unnecessary proximity to a radioactive object.
    \item Incorrect categorisation of waste.
    \item Breaching containment. 
\end{itemize}
Even in the case of `universal' errors, the context of the scenario may affect how severe the error is - and therefore the necessary ways for recovery. For instance, incorrect grasping of a sharp object by a robot may cause only superficial damage to most robot arms. In the Nuclear Glovebox scenario, this can cause a breach in its protective layer, allowing internal components of the robot to be exposed to radiation, or even breach overall containment.

Identifying errors is the subject of much research, especially in human-robot interactions. A dataset with a challenge was created to approach the problem~\cite{spitale2024err}. One potential method that has been employed is to try and detect anomalies in performance~\cite{polenghi2024framework}. This has mostly focused on errors caused by faulty components. 

\subsection{Identifying Causes}

Once an error has been identified, the causes may affect the correct course for recovery, or even if immediate recovery is possible. Identifying causes may also enable further actions to prevent errors at a later stage. Every error may have many causes e.g. in the example of incorrect grasping of a sharp object:
\begin{itemize}
    \item Sharp edge was not identified. Which could in turn have further causes:
    \begin{itemize}
        \item Insufficient Lighting.
        \item Camera view occluded.
    \end{itemize}
    \item Unexpected position drift.
    \begin{itemize}
        \item Servo/motor noise.
        \item Servo/motor failure.
        \item Operator error - in case of teleoperation.
    \end{itemize}
    \item Unexpected obstruction
    \begin{itemize}
        \item Insufficient lighting
        \item Camera view occluded
    \end{itemize}
    \item Human Error
\end{itemize}
In our case of gloveboxes, they tend to suffer from poor visibility due to small viewing hatches or clouded surfaces. As the environment has limited visibility for both human operators and cameras placed either inside or outside the glovebox, collisions and misidentifications of objects can occur. Such lack of information cannot be solved by simply adding more cameras, and needs to be managed during operation, e.g. get visuals from different angles of a non-clouded area, perform operator practices to measure distances during reduced visibility. 

In the case of something like unexpected position drift, a small course correction could be a method of immediate recovery, especially if it has only been caused by noise in the sensors/motor. However, if a motor failure has occurred, this may require the robot to be repaired or replaced.

\subsection{Communicating Errors}

It is important to communicate errors once they occur. Even when a robot is operating autonomously, errors are also a process by which we learn~\cite{mera2022unraveling}. Communicating what caused an error allows for processes to be improved. In collaborating teams, it is even more important to recognise and communicate errors. Recovery from an error may take time, or cause an autonomous system to behave unexpectedly when interacting with other agents. The agent that detects the error may also not be the one that causes the error, or is able to recover, and this may require action from other agents in the team. Errors may also appear in communication itself, which must be resolved.

Errors themselves can occur rapidly, and human response to them is often quick. Previous work has shown that being prepared to rapidly repair communication can be better than trying to be entirely accurate initially~\cite{wallbridge2019generating, wallbridge2021effectiveness}.

There are many ways to communicate errors, and aspects such as the type of error and the goals of the system can influence which is the best way to communicate:
\begin{itemize}
    \item Logs - This can be useful for overall improvements to a system, reducing errors in the long run.
    \item Speech - Direct communication between participating agents.
    \item GUI Elements - Using warning messages on a screen.
    \item Haptic Feedback - Potentially useful in direct recovery during teleoperation. 
\end{itemize}

Communication should be both ways. Humans are very adept at spotting mistakes and correcting for them. Potentially, one of the best ways to enable a system to adapt to mistakes is being able to respond rapidly when a human identifies an error. While often errors are thought of as something only to avoid, they are also one of the best methods of learning. 

\section{Designing a System}

Based on these challenges, we can see that a robust error detection and recovery system can be set into several components (see fig~\ref{fig:simple_detection}):
\begin{figure}[ht]
    \centering
    \includegraphics[width=\columnwidth]{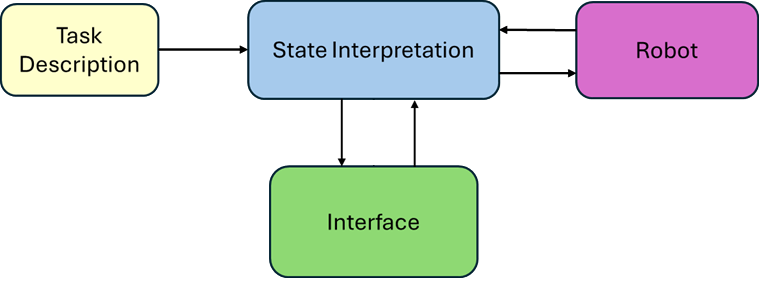}
    \caption[Diagram showing the major components of error recovery]{Diagram showing major components for error detection and recovery. Task description and the robot feed into state interpretation. State interpretation informs the interface, and the robot.}
    \label{fig:simple_detection}
\end{figure}
\begin{description}
    \item[Task Description] - In order to determine errors, the task must be defined. Depending on the task, this might be a single description used at the beginning of the task, or it may be a constantly updating flow.
    \item[State Interpretation] - This needs to take into account several factors - the goal, as defined by the task description. It will also need to define the errors, in part based on the context, as well as the actual task.
    \item[Robot] - The robot itself will need to provide sensor information to allow for state interpretation to be formed, as well as its own state. In turn, it must also be prepared to act on information from State Interpretation, to react to an error to recover.
    \item[Interface] - The system needs to inform users/operators of errors, to allow for both recovery in teams and long-term improvements. The interface especially can benefit from participatory design. It is important to consider how errors are reported, especially in collaborative teams. Too many reports or false flags can lead to people ignoring automated reports. There also needs to be the ability for this interface to work both ways - humans may be the best source of error reporting in collaborative teams, and may enable a system to learn to identify errors more successfully. 
\end{description}

\section{Conclusion}

In this paper, we have looked at why error recovery is important -it can vastly increase the reliability of a system, especially in complex and real-world environments. While progress is being made, many challenges still remain to allow an autonomous system -especially one that collaborates with people. Such a system needs to ensure to take in the context from the task, as well as communicate between the interface and robot to interpret the current state, and what needs to be done.

\section{Acknowledgements}
We acknowledge the resources, funding and support provided by the Robotics and AI Collaboration (RAICo) that has made this work possible.

\bibliographystyle{ACM-Reference-Format}
\bibliography{lit}

@article{russakovsky2015imagenet,
  title={Imagenet large scale visual recognition challenge},
  author={Russakovsky, Olga and Deng, Jia and Su, Hao and Krause, Jonathan and Satheesh, Sanjeev and Ma, Sean and Huang, Zhiheng and Karpathy, Andrej and Khosla, Aditya and Bernstein, Michael and others},
  journal={International journal of computer vision},
  volume={115},
  number={3},
  pages={211--252},
  year={2015},
  publisher={Springer}
}

@inproceedings{zellers2019hellaswag,
  title={Hellaswag: Can a machine really finish your sentence?},
  author={Zellers, Rowan and Holtzman, Ari and Bisk, Yonatan and Farhadi, Ali and Choi, Yejin},
  booktitle={Proceedings of the 57th annual meeting of the association for computational linguistics},
  pages={4791--4800},
  year={2019}
}

@book{moravec1988mind,
  title={Mind children: The future of robot and human intelligence},
  author={Moravec, Hans},
  year={1988},
  publisher={Harvard University Press}
}

@article{ni2025survey,
  title={A survey on large language model benchmarks},
  author={Ni, Shiwen and Chen, Guhong and Li, Shuaimin and Chen, Xuanang and Li, Siyi and Wang, Bingli and Wang, Qiyao and Wang, Xingjian and Zhang, Yifan and Fan, Liyang and others},
  journal={arXiv preprint arXiv:2508.15361},
  year={2025}
}

@article{mera2022unraveling,
  title={Unraveling the benefits of experiencing errors during learning: Definition, modulating factors, and explanatory theories},
  author={Mera, Yeray and Rodr{\'\i}guez, Gabriel and Marin-Garcia, Eugenia},
  journal={Psychonomic bulletin \& review},
  volume={29},
  number={3},
  pages={753--765},
  year={2022},
  publisher={Springer}
}

@article{jang2014empirical,
  title={An empirical study on the human error recovery failure probability when using soft controls in NPP advanced MCRs},
  author={Jang, Inseok and Kim, Ar Ryum and Jung, Wondea and Seong, Poong Hyun},
  journal={Annals of Nuclear Energy},
  volume={73},
  pages={373--381},
  year={2014},
  publisher={Elsevier}
}

@article{reason2000human,
  title={Human error: models and management},
  author={Reason, James},
  journal={Bmj},
  volume={320},
  number={7237},
  pages={768--770},
  year={2000},
  publisher={British Medical Journal Publishing Group}
}

@techreport{dennison1995application,
  title={Application of glove box robotics to hazardous waste management},
  author={Dennison, David K and Hurd, Randall L and Merrill, Roy D and Reitz, Thomas C},
  year={1995},
  institution={Lawrence Livermore National Lab., CA (United States)}
}

@article{wallbridge2021effectiveness,
  title={The effectiveness of dynamically processed incremental descriptions in human robot interaction},
  author={Wallbridge, Christopher D and Smith, Alex and Giuliani, Manuel and Melhuish, Chris and Belpaeme, Tony and Lemaignan, S{\'e}verin},
  journal={ACM Transactions on Human-Robot Interaction (THRI)},
  volume={11},
  number={1},
  pages={1--24},
  year={2021},
  publisher={ACM New York, NY}
}

@article{wallbridge2019generating,
  title={Generating spatial referring expressions in a social robot: Dynamic vs. non-ambiguous},
  author={Wallbridge, Christopher D and Lemaignan, S{\'e}verin and Senft, Emmanuel and Belpaeme, Tony},
  journal={Frontiers in Robotics and AI},
  volume={6},
  pages={67},
  year={2019},
  publisher={Frontiers Media SA}
}

@article{li2024autoflow,
  title={Autoflow: Automated workflow generation for large language model agents},
  author={Li, Zelong and Xu, Shuyuan and Mei, Kai and Hua, Wenyue and Rama, Balaji and Raheja, Om and Wang, Hao and Zhu, He and Zhang, Yongfeng},
  journal={arXiv preprint arXiv:2407.12821},
  year={2024}
}

@article{xu2024hallucination,
  title={Hallucination is inevitable: An innate limitation of large language models},
  author={Xu, Ziwei and Jain, Sanjay and Kankanhalli, Mohan},
  journal={arXiv preprint arXiv:2401.11817},
  year={2024}
}

@inproceedings{spitale2024err,
  title={Err@ hri 2024 challenge: Multimodal detection of errors and failures in human-robot interactions},
  author={Spitale, Micol and Parreira, Maria Teresa and Stiber, Maia and Axelsson, Minja and Kara, Neval and Kankariya, Garima and Huang, Chien-Ming and Jung, Malte and Ju, Wendy and Gunes, Hatice},
  booktitle={Proceedings of the 26th International Conference on Multimodal Interaction},
  pages={652--656},
  year={2024}
}

@article{polenghi2024framework,
  title={A framework for fault detection and diagnostics of articulated collaborative robots based on hybrid series modelling of Artificial Intelligence algorithms},
  author={Polenghi, Adalberto and Cattaneo, Laura and Macchi, Marco},
  journal={Journal of Intelligent Manufacturing},
  volume={35},
  number={5},
  pages={1929--1947},
  year={2024},
  publisher={Springer}
}

@misc{rrobo,
  author       = {BBC, Sellafield},
  title        = {Robotic arms could aid nuclear 'glovebox' clean-up},
  year         = {2026},
  url          = {https://www.bbc.co.uk/news/articles/cd979xn2v0go},
  note         = {Accessed: 2026-02-27}
}

@misc{robox,
  author       = {RAICo, UKAEA},
  title        = {RAICo deployments},
  year         = {2025},
  url          = {https://raico.org/technology/deployments/},
  note         = {Accessed: 2026-02-27}
}

\end{document}